\documentclass[]{article}

\usepackage{rotating}
\usepackage{xcolor}
\usepackage{soul}
\usepackage[utf8]{inputenc}
\usepackage[small]{caption}
\usepackage{textcomp}
\usepackage{enumitem} %description list bold label
\usepackage{graphicx}
\usepackage{multirow}
\usepackage{array}
\usepackage{caption}
\usepackage[draft]{hyperref}
\usepackage{balance} % to better equalize the last page
\usepackage{graphics} % for EPS, load graphicx instead
 \usepackage{times} % comment if you want LaTeX's default font
\usepackage{url}  % llt: nicely formatted URLs
\usepackage{color}
\usepackage{verbatim}
\usepackage{multirow}
\usepackage{amssymb}
\usepackage[lined,boxed,linesnumbered,ruled]{algorithm2e}
\usepackage{epstopdf}
\usepackage[]{algorithm2e}
\usepackage{textcomp}

\usepackage{graphicx}
\usepackage{caption}
\usepackage{subcaption}
\usepackage{makecell}
\usepackage{etoolbox}
\usepackage{epstopdf}
\usepackage{graphicx}
\usepackage{color} 
\usepackage{verbatim}
\usepackage{multirow}
\usepackage{enumitem}
\usepackage[lined,boxed,linesnumbered,ruled]{algorithm2e}
\usepackage{epsfig}
\usepackage{commath} 
\usepackage{xcolor,colortbl}
\usepackage{caption}
\usepackage{subcaption}
\usepackage{makecell}
\usepackage{latexsym} 
\usepackage{mathtools}

\newtheorem{myproof}{Proof}
\newtheorem{mylemma}{Lemma}
\newtheorem{mydefinition}{Definition}

\newtheorem{mytheorem}{Theorem}

\def \eg{{\em e.g.,}}
\def \ie{{\em i.e.,}}

\def \one{{\em i)}}
\def \two{{\em ii)}}
\def \three{{\em iii)}}

\newcommand{\obj}{o}
\newcommand{\of}{w}

\title{Reachability Analysis of Deep Neural Networks with Provable Guarantees\footnote{This is the long version of the conference paper accepted in IJCAI-2018, see~\cite{RHK2018}.}}

\author{
Wenjie Ruan$^1$, 
Xiaowei Huang$^2$, 
Marta Kwiatkowska$^1$
\\ 
$^1$ Department of Computer Science, University of Oxford, UK \\
\{wenjie.ruan; marta.kwiatkowska\}@cs.ox.ac.uk\\
$^2$ Department of Computer Science, University of Liverpool, UK\\
xiaowei.huang@liverpool.ac.uk
}
\date{}

\begin{document}

\maketitle

\begin{abstract}
Verifying correctness of deep neural networks (DNNs) is challenging. We study a generic reachability problem for feed-forward DNNs which, for a given set of inputs to the network and a Lipschitz-continuous function over its outputs, computes the lower and upper bound on the function values. Because the network and the function are Lipschitz continuous, all values in the interval between the lower and upper bound are reachable. We show how to obtain the safety verification problem, the output range analysis problem and a robustness measure by instantiating the reachability problem. We present a novel algorithm based on adaptive nested optimisation to solve the reachability problem. The technique has been implemented and evaluated on a range of DNNs, demonstrating its efficiency, scalability and ability to handle a broader class of networks than state-of-the-art verification approaches.

\end{abstract}
		%\vspace{-1mm}

%\vspace{-3pt}
\section{Introduction}
%\vspace{-1pt}

Concerns have been raised about the suitability of deep neural networks (DNNs), or systems with DNN components, for deployment in safety-critical applications, see e.g., \cite{AOSCSM2016,SWRHKK2018}. To ease this concern and gain users' trust, DNNs need to be certified similarly to %software and hardware 
systems such as airplanes and automobiles. In this paper, we propose to study a generic reachability problem which, for a given DNN, an input subspace and a function over the outputs of the network, computes the upper and lower bounds over the values of the function. The function is generic, with the only requirement that it is Lipschitz continuous. %and has the DNN's output as its input. 
We argue that this problem is fundamental for certification of DNNs, %can be a core problem towards certifying DNNs 
as it can be instantiated into several key correctness problems, including adversarial example generation \cite{szegedy2014intriguing,GSS2014}, safety verification \cite{HKWW2017,katz2017reluplex,RWSHKK2018}, output range analysis \cite{LM2017,dutta2017output}, and robustness comparison.

%$R_j(X')$ for a given class label $j\in [1..m]$.
% is the set of possible confidence levels $c_j(x)$ over the label $j$. 
% An error bound may be needed for practical reasons. 
To certify a system, a certification approach needs to provide not only a result but also a guarantee over the result, such as the error bounds. 
Existing approaches for analysing DNNs with a guarantee work by either reducing the problem to a constraint satisfaction problem that can be solved by MILP \cite{LM2017,CNR2017,bunel2017piecewise,xiang2017output}, SAT \cite{NKPSW2017} or SMT \cite{katz2017reluplex,bunel2017piecewise} techniques, or applying search algorithms over discretised vector spaces \cite{HKWW2017,WHK2017}. Even though they are able to achieve guarantees, they suffer from two major weaknesses. Firstly, their subjects of study are restricted. 
More specifically, they can only work with layers conducting linear transformations (such as convolutional and fully-connected layers) and simple non-linear transformations (such as ReLU), and cannot work with other important layers, such as the sigmoid, max pooling and softmax layers that are widely used in state-of-the-art networks. Secondly, the scalability of the constraint-based approaches is significantly limited by both the capability of the solvers and the size of the network, and they can only work with networks with a few hundreds of hidden neurons. However, state-of-the-art networks usually have millions, or even billions, of hidden neurons. 

This paper proposes a novel approach to tackle the generic reachability problem, which does not suffer from the above weaknesses and provides provable guarantees %over its results 
in terms of the upper and lower bounds over the errors. The approach is inspired by recent advances made in the area of global optimisation~\cite{gergel2016adaptive,grishagin2018convergence}. 
For the input subspace defined over a set of input dimensions, an adaptive nested optimisation algorithm is developed. The performance of our algorithm is not dependent on the size of the network and it can therefore scale to work with large networks. 

Our algorithm assumes certain knowledge about the DNN. However, instead of directly translating the activation functions and their parameters (i.e., weights and bias) into linear constraints, % as done in the reduction approaches, 
it needs a Lipschitz constant of the network. For this, we show that several layers that cannot be directly translated into linear constraints are actually Lipschitz continuous, and we are able to compute a tight Lipschitz constant by analysing the activation functions and their parameters. 

We develop a software tool DeepGO\footnote{Available on \url{https://github.com/trustAI/DeepGO}.} and evaluate its performance by comparing with existing constraint-based approaches, namely, SHERLOCK \cite{dutta2017output} and Reluplex \cite{katz2017reluplex}. We also demonstrate % on several problems that can be handled by them. Our experiments include those networks that cannot be handled by existing algorithms. 
our tool on DNNs that are beyond the capability of existing tools.

%Practically, the layer-to-layer mapping functions can be various types of activation functions such as ReLU, Leaky ReLU, sigmoid function, Hyperbolic tangent function etc., or different pooling operations such as max pooling and contrast-normalization pooling, and softmax layer (usually in the last layer). Noted that existing methods of reachability analysis for DNNs are all based on MLP [?], SAT [?] or SMT [?] techniques, which enable them has the following three major limitations: \one~they are only workable on the ReLU activation functions due to the linearizion formulation; \two~they cannot analyses softmax layer which enables their reachability analysis not complete; and \three~those methods are based on a layer-by-layer analysis which makes them only workable for a small-scale neural network(such as 3-layer NN with only hundreds of neurons. 

%To deal with the above limitation, in this paper, we tackle the problem from a "black-box" point of view, which is neither depends on the layer-by-layer analysis nor built upon any MLP solvers. The only assumption we rely on is that the targeted deep neural network satisfies the Lipschitz continuity assumption.
		%\vspace{1mm}

%\vspace{-3pt}
\section{Related Works}
%\vspace{-1pt}

We discuss several threads of work concerning problems that can be obtained by instantiating our generic reachability problem. Their instantiations are explained in the paper. Due to space limitations, this review is by no means complete. 

\vspace{2mm}
\noindent{\bf Safety Verification}
There are two ways of achieving safety verification for DNNs. The first is to reduce the problem into a constraint solving problem. Notable works include, e.g., \cite{PT2010,katz2017reluplex}. However, %as stated, 
they can only work with small networks with hundreds of hidden neurons. The second is to discretise the vector spaces of the input or hidden layers and then apply exhaustive search algorithms or Monte Carlo tree search algorithm on the discretised spaces. The guarantees are achieved by establishing local assumptions such as minimality of manipulations in \cite{HKWW2017} and minimum confidence gap for Lipschitz networks in \cite{WHK2017}. 

\vspace{2mm}
\noindent{\bf Adversarial Example Generation}
Most existing works, e.g., \cite{szegedy2014intriguing,GSS2014,AJJ2014,SAOP2016,CW2016}, apply various heuristic algorithms, %with most of them 
generally using search algorithms based on gradient descent or evolutionary techniques. \cite{PMJFCS2015} construct a saliency map of the importance of the pixels based on gradient descent and then modify the pixels. In contrast with our approach based on global optimisation and works on safety verification, these methods may be able to find adversarial examples efficiently, but are \emph{not able to conclude the nonexistence} of adversarial examples when the algorithm fails to find one.

\vspace{2mm}
\noindent{\bf Output Range Analysis}
The safety verification approach can be adapted to work on this problem. Moreover, \cite{LM2017} consider %the problem of 
determining whether an output value of a DNN is reachable from a given input subspace, and propose an MILP solution. %reduce the problem to a MILP problem. 
\cite{dutta2017output} study the range of output values from a given input subspace. Their method interleaves local search (based on gradient descent) with global search (based on reduction to MILP). Both approaches can only work with small networks. 

		%\vspace{-1mm}

%\vspace{-3pt}
\section{Lipschitz Continuity of DNNs}\label{sec:lipschitz}
%\vspace{-1pt}

This section shows that feed-forward DNNs are Lipschitz continuous.
%and presents an approach to compute, for a given DNN, its Lipschitz constant. 
%
%preliminaries and mathematical problem formulations.
%
%\subsection{Preliminaries}
%
%
%Let $N$ be a network with a set $C$ of classes. Given an input $\inputImage$ and a class $c \in C$, we use $N(\inputImage,c)$ to denote the confidence (expressed as a probability value obtained from normalising the score) of $N$ believing that $\inputImage$ is in class $c$. Moreover, we write $N(\inputImage) = \arg\max_{c\in C} N(\inputImage,c)$ for the class into which $N$ classifies $\inputImage$. 
%%z_Allsorted(1,end)
%For our discussion of image classification networks, the input domain $\inputdomain$ is a vector space, which in most cases can be represented as 
%%the pixel domain (i.e. all possible values a single pixel in an image can take on). For many images, the pixel domain is an alias for 
%${\rm I\!R_{[0,255]}^{w\times h\times ch}}$, where $w,h,ch$ are the width, height, and number of channels of an image, respectively, and we let $P_0 = w\times h\times ch$ be the set of input dimensions. In the following, we may refer to an element in $w\times h$ as a pixel and an element in $P_0$ as a dimension. 
%
%This paper focuses on the confidence reachability analysis of a multi-layer feed-forward neural networks. 
Let $f: \mathbb{R}^n \rightarrow \mathbb{R}^m$ be a $N$-layer 
%feed-forward neural 
network such that, for a given input $x\in \mathbb{R}^n$, $f(x) = \{c_1,c_2,...,c_m\}\in \mathbb{R}^m$ represents the confidence values for $m$ classification labels. Specifically, we have 
$	f(x) = f_N(f_{N-1}(...f_1(x;W_1,b_1);W_2,b_2);...);W_N,b_N)
$ where $W_i$ and $b_i$ for $i = 1,2,...,N$ are learnable parameters and $f_i(z_{i-1};W_{i-1},b_{i-1})$ is the function mapping from the output of layer $i-1$ to the output of layer $i$ such that $z_{i-1}$ is the output of layer $i-1$. Without loss of generality, we normalise the input to lie $x\in [0,1]^n$. The output $f(x)$ is usually normalised to be in $[0,1]^m$ with a softmax layer.

\begin{mydefinition}[Lipschitz Continuity]
	Given two metric spaces $(X, d_X)$ and $(Y, d_Y)$, where $d_X$ and $d_Y$ are the metrics on the sets $X$ and $Y$ respectively, a function $f: X\rightarrow Y$ is called {\em Lipschitz continuous} if there exists a real constant $K\geq0$ such that, for all $x_1, x_2 \in X$:
	\begin{equation}
	 d_Y(f(x_1), f(x_2)) \le K d_X(x_1, x_2).
	\end{equation}
	$K$ is called the {\em Lipschitz constant} for the function $f$. The smallest $K$ is called {\em the Best Lipschitz constant}, denoted as $K_{best}$.
\end{mydefinition}

In \cite{szegedy2014intriguing}, the authors show that deep neural networks with half-rectified layers (\ie~convolutional or fully connected layers with ReLU activation functions), max pooling and contrast-normalization layers are Lipschitz continuous. They prove that the upper bound of the Lipschitz constant can be estimated via the operator norm of learned parameters $W$.

Next, we show that the softmax layer, sigmoid and Hyperbolic tangent activation functions also satisfy Lipschitz continuity. First we need the following lemma~\cite{sohrab2003basic}.

\begin{mylemma}\label{theo-1}
Let $f: \mathbb{R}^n \rightarrow \mathbb{R}^m$,	if $||\partial{f(x)}/\partial{x}|| \leq K$ for all $x \in [a, b]^n$, then $f$ is Lipschitz continuous on $[a, b]^n$ and $K$ is its Lipschitz constant, where $||*||$ represents a norm operator.
\end{mylemma}

Based on this lemma, we have the following theorem.

\begin{mytheorem}
Convolutional or fully connected layers with the sigmoid activation function $s(Wx+b)$, Hyperbolic tangent activation function $t(Wx+b)$, and softmax function $p(x)_j$ are Lipschitz continuous and their Lipschitz constants are $\dfrac{1}{2}\norm{W}$, $\norm{W}$, and $\sup_{i,j}(\norm{x_i} + \norm{x_ix_j})$, respectively.
\end{mytheorem}

\begin{myproof}
First of all, we show that the norm operators of their Jacobian matrices are bounded.

(1)~Layer with sigmoid activation $s(q) = 1/(1+e^{-q})$ with $ q = Wx+b$:
\begin{equation}
\begin{split}
\norm{\dfrac{\partial{s(x)}}{\partial{x}}} = \norm{\dfrac{\partial{s(q)}}{\partial{q}} \dfrac{\partial{q}}{\partial{x}}}
\leq \norm{\dfrac{\partial{s(q)}}{\partial{q}}} \norm{\dfrac{\partial{q}}{\partial{x}}}\\
\leq \norm{s(q)\circ(\mathbf{1}-s(q))}\norm{W}\leq \dfrac{1}{2}\norm{W}
\end{split}
\end{equation}

(2)~Layer with Hyperbolic tangent activation function $t(q) = 2/(1+e^{-2q})-1$ with $ q = Wx+b$:
\begin{equation}
\begin{split}
\norm{\dfrac{\partial{t(x)}}{\partial{x}}} = \norm{\dfrac{\partial{t(q)}}{\partial{q}} \dfrac{\partial{q}}{\partial{x}}}
\leq \norm{\dfrac{\partial{t(q)}}{\partial{q}}} \norm{\dfrac{\partial{q}}{\partial{x}}}\\
\leq \norm{\mathbf{1}-t(q)\circ t(q))}\norm{W}\leq \norm{W}
\end{split}
\end{equation}

(3)~Layer with softmax function $p(x)_j = e^{x_j}/(\sum_{k = 1}^{n}{e^{x_k}})$ for $j = 1,...,m$ and $n = m$ (dimensions of input and output of softmax are the same):
\begin{equation}
\begin{split}
\norm{\dfrac{\partial{p(x)_j}}{\partial{x_i}}} = 
\left\{
\begin{array}{ll}
x_i(1-x_j), ~~i = j\\
-x_ix_j,~~i\ne j
\end{array}
\right. \leq \sup_{i,j} (\norm{x_i} + \norm{x_ix_j})
\end{split}
\end{equation}
Since the softmax layer is the last layer of a deep neural network, we can estimate its supremum based on Lipschitz constants of previous layers and box constraints of DNN's input.

The final conclusion follows by Lemma 1 and the fact that all the layer functions are bounded on their Jacobian matrix. 
\end{myproof}

%\vspace{-3pt}
\section{Problem Formulation} 
%\vspace{-1pt}

%In this section, we present the formulate the problem of confidence reachability of a neural network. 
Let $\obj: [0,1]^m \rightarrow \mathbb{R}$ be a Lipschitz continuous function statistically evaluating the outputs of the network. Our problem is to find its upper and lower bounds given the set $X'$ of inputs to the network. Because both the network $f$ and the %statistical evaluation 
function $\obj$ are Lipschitz continuous, all values between the upper and lower bounds have a corresponding input, i.e., are reachable. 

\begin{mydefinition}[Reachability of Neural Network]
Let $X'\subseteq [0,1]^n$ be an input subspace and $f: \mathbb{R}^n \rightarrow \mathbb{R}^m$ a network.
% such that $f(x) = \{c_1,...,c_j,...,c_m\}\in [0,1]^m$ and $c_j$ is the $j$-th confidence output representing the probability of input $x$ belonging to $j$-th classification label. We also use $c_j(x)$ to represent the $j$-th output of DNN given input $x$. Then 
The reachability of $f$ over the function $\obj$ under an error tolerance $\epsilon \geq 0$ is a set $R(\obj,X',\epsilon) = [l,u]$ such that 
\begin{equation}
%R(\bar{x}) = \{[a,b] | \max\limits_{1\leq j\leq n}\}
%R_j(X') = \{ r | \min \limits_{\bar{x}} c_j(\bar{x}) \leq |r-\epsilon_j| \leq \max \limits_{\bar{x}} c_j(\bar{x}) \text{ for } \bar{x} \in X', r \geq 0 \}.
l \geq \inf_{x' \in X'} \obj(f(x')) -\epsilon \text { and } u \leq \sup_{x' \in X'} \obj(f(x')) + \epsilon.
\end{equation}
We write $u(\obj,X',\epsilon)=u$ and $l(\obj,X',\epsilon)=l$ for the upper and lower bound, respectively. 
%Confidence Upper Bound (CUB) and Confidence Lower Bound (CLB), respectively. 
Then the reachability diameter is 
	\begin{equation}
           D(\obj,X',\epsilon) =u(\obj,X',\epsilon)-l(\obj,X',\epsilon).
	\end{equation}
Assuming these notations, we may write $D(\obj,X',\epsilon;f)$ if we need to explicitly refer to the network $f$. 
\end{mydefinition}

In the following, we instantiate $\obj$ with a few concrete functions, and show that several key verification problems for DNNs can be reduced to our reachability problem. 

\begin{mydefinition}[Output Range Analysis]
Given a class label $j\in [1,..,m]$, we let $\obj = \Pi_j$ such that $\Pi_j((c_1,...,c_m))=c_j$. 
\end{mydefinition}

We write $c_j(x) = \Pi_j(f(x))$ for the network's confidence in classifying $x$ as label $j$. 
Intuitively, output range \cite{dutta2017output} quantifies how a certain output of a deep neural network (\ie~classification probability of a certain label $j$) varies in response to a set of DNN inputs with an error tolerance $\epsilon$. Output range analysis can be easily generalised to logit \footnote{Logit output is the output of the layer before the softmax layer. The study of logit outputs is conducted in, e.g., \cite{PMJFCS2015,dutta2017output}.} range analysis.

We show that the safety verification problem \cite{HKWW2017} can be reduced to solving the reachability problem. 

\newcommand{\Safe}{{\tt S}}
\newcommand{\Robust}{{\tt R}}

\begin{mydefinition}[Safety]
A network $f$ is safe with respect to an input $x$ and an input subspace $X'\subseteq [0,1]^n$ with $x \in X'$, written as $\Safe(f,x,X')$, if 
\begin{equation}
\forall x' \in X': \arg\max_{j} c_j(x') = \arg\max_{j} c_j(x)
\end{equation} 
%where $c_j(x) = f(x)_j$ returns $N$'s confidence in classifying $x$ as label $j$. 
\end{mydefinition}

We have the following reduction theorem. 

%For the reduction, we define two functions $o_1=\Pi_j$ such that $j=\arg\max_{j}c_j(x)$, and $o_2 = +_{-j}$ such that $+_{-j}(c_1,...,c_m) = \sum_{i=1, i\neq j}^{m} c_i$. Therefore, we have 

\begin{mytheorem}\label{thm:safety}
A network $f$ is safe with respect to $x$ and $X'$ s.t. $x \in X'$ if and only if
$u(\oplus,X',\epsilon) \leq 0$,
%$l(\Pi_j,X',\epsilon) \geq u(\oplus_{-j},X',\epsilon)$
where $\oplus(c_1,...,c_m) = \max_{i\in \{1..m\}}(\Pi_{i} (c_1,...,c_m) - \Pi_{j} (c_1,...,c_m))$ and $j=\arg\max_{j}c_j(x)$. The error bound of the safety decision problem by this reduction is $2\epsilon$. 
\end{mytheorem}

It is not hard to see that the adversarial example generation \cite{szegedy2014intriguing}, which is to find an input $x' \in X'$ such that $\arg\max_{j} c_j(x') \neq \arg\max_{j} c_j(x)$, is the dual problem of the safety problem. 

The following two problems define the robustness comparisons between the networks and/or the inputs. 

\begin{mydefinition}[Robustness]
Given two homogeneous\footnote{ Here, two networks are homogeneous if they are applied on the same classification task but may have different network architectures (layer numbers, layer types, etc) and/or parameters.}
%E.g., two different neural networks but both are for the MNIST image classification task.} 
networks $f$ and $g$, we say that $f$ is strictly more robust than $g$ with respect to a function $\obj$, an input subspace $X'$ and an error bound $\epsilon$, written as $\Robust_{o,X',\epsilon}(f,g)$, if $D(\obj,X',\epsilon;f) < D(\obj,X',\epsilon;g)$.

%$j$-th output and an input subspace $\bar{x}$ if $D_j(\bar{x};f(x)) < D_j(\bar{x};g(x))$.
\end{mydefinition}

\begin{mydefinition}
	Given two input subspaces $X'$ and $X''$ and a network $f$, we say that $f$ is more robust on $X'$ than on $X''$ with respect to a statistical function $\obj$ and an error bound $\epsilon$, written as $\Robust_{f,o,\epsilon}(X',X'')$, if $D(\obj,X',\epsilon) < D(\obj,X'',\epsilon)$.
%$D_j(\bar{x}) < D_j(\hat{x})$.
\end{mydefinition}

Thus, by instantiating the function $\obj$, we can quantify the output/logit range of a network, evaluate whether a network is safe, and compare the robustness of two homogeneous networks or two input subspaces for a given network.
		%\vspace{-1mm}

%\vspace{-3pt}
\section{Confidence Reachability with Guarantees}
%\vspace{-1pt}

Section~\ref{sec:lipschitz} shows that a trained deep neural network is Lipschitz continuous regardless of its layer depth, activation functions and number of neurons. Now, to solve the reachability problem, we need to find the \emph{global} minimum and maximum values given an input subspace, assuming that we have a Lipschitz constant $K$ for the function $o\!\cdot\!f$. In the following, we let $\of = o\!\cdot\!f$ be the concatenated function. Without loss of generality, we assume the input space $X'$ is a box-constraint, which is clearly feasible since images are usually normalized into $[0,1]^n$ before being fed into a neural network.

% Thus we need to solve the following problem. Here we start to use $x$ instead of $\bar{x}$ (we can treat the fixed dimensions in input space as a learned parameters in neural network) and drop index $j$ (since we can perform the some analysis to each output separately). 
The computation of the minimum value is reduced to solving the following optimization problem with guaranteed convergence to the global minimum (the maximization problem can be transferred into a minimization problem):
\begin{equation}\label{equ-8}
\begin{split}
\min\limits_{x}~~\of(x),~~~s.t.~~x \in [a,b]^n
\end{split}
\end{equation}
However, the above problem is very difficult since $\of(x)$ is a highly non-convex function which cannot be guaranteed to reach the global minimum by regular optimization schemes based on gradient descent. Inspired by an idea from optimisation, see e.g., \cite{piyavskii1972algorithm,TZ1989}, we design another continuous function $h(x,y)$, which serves as a lower bound of the original function $\of(x)$. Specifically, we need 
%This idea is also widely adopted in optimization community.
\begin{equation}\label{equ-9}
\begin{split}
\forall x, y \in [a,b]^n,~ h(x,y) \leq \of(x)~~\text{and}~~h(x,x) = \of(x)
\end{split}
\end{equation}
Furthermore, for $i\geq 0$, we let $\mathcal{Y}_i = \{y_0,y_1,...,y_i\} $ be a finite set containing $i+1$ points from the input space $ [a, b]^n$, and let $\mathcal{Y}_i \subseteq \mathcal{Y}_k$ when $k>i$, then we can define a function
$H(x;\mathcal{Y}_i) = \max_{y\in \mathcal{Y}_i } h(x,y)$
 which satisfies the following relation:
\begin{equation}
H(x; \mathcal{Y}_i) < H(x; \mathcal{Y}_k)\leq %\inf_{x\in [a,b]^n}
\of(x), \forall i < k
\end{equation}

We use $l_i = \inf_{x\in [a,b]^n} H(x;\mathcal{Y}_i)$ to denote the minimum value of $H(x;\mathcal{Y}_i)$ for $x\in [a,b]^n$. Then we have 
%can get the following relation:
\begin{equation}
l_0< l_1< ... < l_{i-1}< l_i \leq \inf_{x\in [a,b]^n} \of(x)
\end{equation}
Similarly, we need a sequence of upper bounds $u_i$ to have 
\begin{equation}\label{eqn-14}
\begin{split}
l_0< ... < l_i \leq \inf_{x\in [a,b]^n} \of(x)\leq u_i< ...< u_0
\end{split}
\end{equation}
By Expression (\ref{eqn-14}), we can have the following:
\begin{equation}\label{eqn-15}
\begin{split}
\lim\limits_{i\to \infty}l_i = \min_{x\in [a,b]^n} \of(x) \text{   and   }
\lim\limits_{i\to \infty}(u_i - l_i) = 0
\end{split}
\end{equation}

Therefore, we can asymptotically approach the global minimum. Practically, we execute a finite number of iterations by using an error tolerance $\epsilon$ to control the termination. 
%
%\begin{myremarks}
%	Whenever terminated, it returns a lower bound and a upper bound of the global minimum, and we can evaluate the quality of global minimum obtained so far.
%\end{myremarks}
%
In next sections, we present our approach, which constructs a sequence of lower and upper bounds, and show that it can converge with an error bound. To handle the high-dimensionality of DNNs, our approach is inspired by the idea of adaptive nested optimisation in \cite{gergel2016adaptive}, with significant differences in the detailed algorithm and convergence proof. 

%We first discuss the one-dimension case and then the multiple-dimension case.
\begin{figure}[t]
	\centering
	\includegraphics[width=1\linewidth]{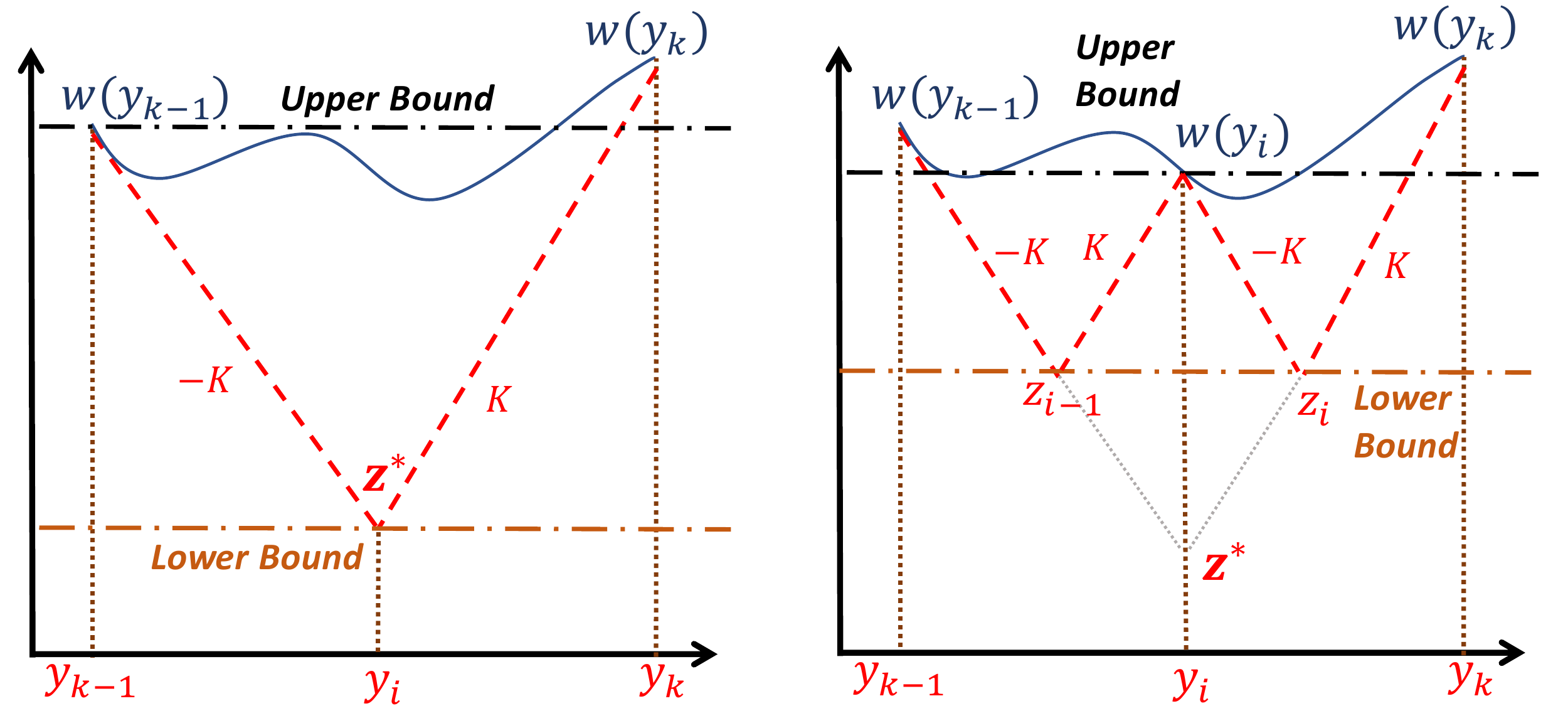}
	\caption{A lower-bound function designed via Lipschitz constant}
	\label{fig-1}
\end{figure}

\subsection{One-dimensional Case}\label{sec:onedimensional}

We first introduce an algorithm which works over one dimension of the input, and therefore is able to handle the case of $x \in [a,b]$ in Eqn.~(\ref{equ-8}).
%, based on Section 2 and Theorem~1, we can estimate the Lipschitz constants of a neural network numerically either or analytically. 
The multi-dimensional optimisation algorithm will be discussed in Section~\ref{sec:multi} by utilising the one-dimensional algorithm. 

We define the following lower-bound function.
\begin{equation}\label{eqn-10}
\begin{split}
h(x,y) = \of(y) - K|x-y| \\
H(x;\mathcal{Y}_i) = \max\limits_{y\in \mathcal{Y}_i }~~\of(y) - K|x-y|
\end{split}
\end{equation}
where $K > K_{best}$ is a Lipschitz constant of $\of$ and $H(x;\mathcal{Y}_i)$ intuitively represents the lower-bound sawtooth function shown as Figure~1.
%$L \geq K_{best}$ and we can estimate by $L = |d(c(x))/dx| + \eta$ $(\eta \geq 0)$ based on Theorem~1. 
The set of points $\mathcal{Y}_i$ is constructed recursively. Assuming that, after $(i-1)$-th iteration, we have $\mathcal{Y}_{i-1} = \{y_0,y_1,..,y_{i-1}\}$, whose elements are in ascending order, and sets 
$$
\of(\mathcal{Y}_{i-1} )=\{\of(y_0), \of(y_1),..,\of(y_{i-1})\}
$$
$$\mathcal{L}_{i-1} = \{l_0,l_1,...,l_{i-1}\}$$
$$\mathcal{U}_{i-1} = \{u_0,u_1,...,u_{i-1}\}$$
$$\mathcal{Z}_{i-1} = \{z_1,...,z_{i-1}\}$$
%where $z_{k} = \dfrac{\of(y_{k})+ \of(y_{k-1})}{2} - \dfrac{K(y_{k} - y_{k-1})}{2}, \forall k = 1,2,..,i-1$. 
%
The elements in sets $\of(\mathcal{Y}_{i-1} )$, $\mathcal{L}_{i-1}$ and $\mathcal{U}_{i-1}$ have been defined earlier. The set $\mathcal{Z}_{i-1}$ records the smallest values $z_k$ computed in an interval $[y_{k-1},y_k]$.

In $i$-th iteration, we do the following sequentially:
\begin{itemize}

%\item Numerically estimate Lipschitz constant as $K = \eta \max_{j = 1,...,i-1} \{(\of(y_j) - \of(y_{j-1}))/(y_j - y_{j-1})\}$ where $\eta > 1$.

	\item Compute $y_i = \arg\inf_{x\in [a,b]} H(x;\mathcal{Y}_{i-1})$ as follows. Let $z^* = \min{\mathcal{Z}_{i-1} }$ and $k$ be the index of the interval $[y_{k-1},y_k]$ where $z^*$ is computed. Then we let 
\begin{equation}\label{equ:newpoint}
y_i = \dfrac{y_{k-1}+y_k}{2} - \dfrac{\of(y_{k}) - \of(y_{k-1})}{2K}
\end{equation}

 and have that $y_i \in (y_{k-1},y_k)$. 
	
	\item Let $\mathcal{Y}_i = \mathcal{Y}_{i-1}\cup\{y_i\}$, then reorder $\mathcal{Y}_i$ in ascending order, and update $\of(\mathcal{Y}_{i} )=\of(\mathcal{Y}_{i-1})\cup \{\of(y_i)\}$.
	
	\item Calculate 
	\begin{equation}\label{eqn-16}
	z_{i-1} = \dfrac{\of(y_{i})+ \of(y_{k-1})}{2} - \dfrac{K(y_{i} - y_{k-1})}{2}\end{equation}
	
	\begin{equation}\label{eqn-17}
	z_{i} = \dfrac{\of(y_{k})+ \of(y_{i})}{2} - \dfrac{K(y_{k} - y_{i})}{2}\end{equation} and update $\mathcal{Z}_{i} = (\mathcal{Z}_{i-1}\setminus \{z^*\}) \cup\{z_{i-1},z_{i}\} $.
	
	\item Calculate the new lower bound $l_i = \inf_{x\in [a,b]} H(x;\mathcal{Y}_i)$ by letting $l_i =\min\mathcal{Z}_{i}$,
%\min \{l_{i-1}, z_{i-1},z_{i}\}$, 
and updating $\mathcal{L}_{i} = \mathcal{L}_{i-1} \cup \{l_i \}$.
	
	\item Calculate the new upper bound $u_i = \min_{y\in \mathcal{Y}_i }\of(y)$ by letting $u_i = \min \{u_{i-1}, \of(y_i)\}$.
\end{itemize}

We terminate the iteration whenever $|u_i-l_i|\leq\epsilon$,
% where $\epsilon$ is the error tolerance, 
and let the global minimum value be $y^* = \min_{x\in [a,b]} H(x;\mathcal{Y}_{i})$ and the minimum objective function be $\of^* = \of(y^*)$. 

Intuitively, as shown in Fig.~1, we iteratively generate lower bounds (by selecting in each iteration the lowest point in the saw-tooth function in the figure) by continuously refining a piecewise-linear lower bound function, which is guaranteed to below the original function due to Lipschitz continuity. The upper bound is the lowest evaluation value of the original function so far. 

\subsubsection{Convergence Analysis}
In the following, we show the convergence of this algorithm to the global minimum by proving the following conditions. 
\begin{itemize}
    \item Convergence Condition 1: $\lim\limits_{i\to \infty}l_i = \min\limits_{x\in [a,b]} \of(x)$
    \item Convergence Condition 2: $\lim_{i\to \infty}(u_i - l_i) = 0$
\end{itemize}

\begin{myproof}[Monotonicity of Lower/Upper Bound Sequences]
First, we prove that the lower bound sequence $\mathcal{L}_i$ is strictly monotonic. 
	Because 
	\begin{equation}l_i = \min \mathcal{Z}_i= \min \{(\mathcal{Z}_{i-1}\setminus \{z^*\}) \cup\{z_{i-1},z_{i}\} \} \end{equation} and $l_{i-1} = \min \mathcal{Z}_i$. To show that $l_i > l_{i-1}$, we need to prove $z_{i-1} > z^*$ and $z_{i} > z^*$. By the algorithm, $z^*$ is computed from interval $[y_{k-1}, y_k]$, so we have %
	\begin{equation}\label{eqn-19}
	z^* = \dfrac{\of(y_{k})+ \of(y_{k-1})}{2} - \dfrac{K(y_{k} - y_{k-1})}{2}\end{equation}
	We then have 
	\begin{equation}\label{equ:lowerbound}
	z_{i-1} - z^* = \dfrac{\of(y_i)-\of(y_k) - K(y_i - y_k)}{2}
	\end{equation} 
	Since $y_i < y_{k}$ and $K>K_{best}$, by Lipschitz continuity we have $z_{i-1} > z^* $. 
	Similarly, we can prove $z_{i} > z^* $. Thus $l_i > l_{i-1}$ is guaranteed. 
	
	Second, the monotonicity of upper bounds $u_i$ can be seen from the algorithm, since $u_i $ is updated to $ \min \{u_i, \of(y_i)\}$ in every iteration. 
\label{proof2}
\end{myproof}

\begin{myproof}[Convergence Condition 1]
	~\\
	Since $\mathcal{Y}_{i-1}\subseteq \mathcal{Y}_{i}$, we have $H(x; \mathcal{Y}_{i-1}) \leq H(x; \mathcal{Y}_{i})$. Based on Proof~\ref{proof2}, we also have $l_{i-1}< l_i$. Then since  %And because 
	\begin{equation}l_i = \inf_{x\in [a,b]} H(x; \mathcal{Y}_{i}) \leq \min_{x\in [a,b]}\of(x)\end{equation} 
	the lower bound sequence $\{l_0,l_1,...,l_i\}$ is strictly monotonically increasing and bounded from above by $\min_{x\in [a,b]}\of(x)$. Thus $\lim_{i\to \infty}l_i = \min_{x\in [a,b]} \of(x)$ holds.
\end{myproof}

\begin{myproof}[Convergence Condition 2]
	~\\
	Since $\lim_{i\to \infty}l_i = \min_{x\in [a,b]} \of(x)$, we show $\lim_{i\to \infty}(u_i - l_i) = 0$ by showing that $\lim_{i\to \infty}u_i = \min_{x\in [a,b]} \of(x)$. 
Since $\mathcal{Y}_{i} = \mathcal{Y}_{i-1} \cup\{y_i\}$ and $y_i \in X =[a, b]$, we have $\lim_{i\to \infty} \mathcal{Y}_{i} = X$. Then we have $\lim_{i\to \infty}u_i = \lim_{i\to \infty} \inf_{y\in \mathcal{Y}_i } \of(y) = \inf {\of(X)}$. Since $X = [a,b]$ is a closed interval, we can prove $\lim_{i\to \infty}u_i = \inf {\of(X)} = \min_{x\in [a,b]} \of(x)$. 
\end{myproof}

\subsubsection{Dynamically Improving the Lipschitz Constant}

A Lipschitz constant closer to $K_{best}$ can greatly improve the speed of convergence of the algorithm. We design a practical approach to dynamically update the current Lipschitz constant according to the information obtained from the previous iteration: 
\begin{equation}K = \eta \max_{j = 1,...,i-1} \abs{\dfrac{\of(y_j) - \of(y_{j-1})}{y_j - y_{j-1}}}\end{equation} where $\eta > 1$. We emphasise that, because 
%during the iteration, we can numerically approximate the Lipschitz constant $K$, as we can show that 
$$ \lim_{i \to \infty} \max_{j = 1,...,i-1} \eta\abs{\dfrac{\of(y_j) - \of(y_{j-1})}{y_j - y_{j-1}}} = \eta \sup_{y\in [a,b]} {\dfrac{d\of}{dy}} > K_{best}
$$
this dynamic update does not compromise the convergence.

\subsection{Multi-dimensional Case}\label{sec:multi}

%We 
%use the nested optimization approach, which is widely applied in optimization community~\cite{gergel2016adaptive}. Its core idea is to
The basic idea is to decompose a multi-dimensional optimization problem into a sequence of nested one-dimensional subproblems. Then the minima of those one-dimensional minimization subproblems are back-propagated into the original dimension and the final global minimum is obtained. 
\begin{equation}\label{equ-16}
\min\limits_{x \in [a_i,b_i]^n}~~\of(x) =
\min\limits_{x_1\in [a_1,b_1]}... \min\limits_{x_n\in [a_n,b_n]} \of(x_1,...,x_n)
\end{equation}

We first introduce the definition of $k$-th level subproblem.

\begin{mydefinition}\label{def:kthlevel}
%[$k$-th Level Subproblem]
	The $k$-th level optimization subproblem, written as $\phi_k(x_1,...,x_k)$, is defined as follows: for $1\leq k \leq n-1$,
	$$\phi_k(x_1,...,x_k) = \min_{x_{k+1}\in [a_{k+1},b_{k+1}]} \phi_{k+1}(x_1,...,x_k, x_{k+1}) $$
	and for $k=n$, $$\phi_n(x_1,...,x_n) = \of(x_1,x_2,...,x_n).$$
\end{mydefinition}
Combining Expression (\ref{equ-16}) and Definition~\ref{def:kthlevel}, we have that $$\min_{{x} \in [a_i,b_i]^n}~~\of({x}) = \min_{x_1\in [a_1,b_1]} \phi_1(x_1)$$ which is actually a one-dimensional optimization problem and therefore can be solved by the method in Section~\ref{sec:onedimensional}. 

However, when evaluating the objective function $\phi_1(x_1)$ at $x_1=a_1$, we need to project $a_1$ into the next one-dimensional subproblem $$\min_{x_2\in [a_2,b_2]}\phi_2(a_1,x_2)$$ We recursively perform the projection until we reach the $n$-th level one-dimensional subproblem, $$\min_{x_n\in [a_n,b_n]}\phi_n(a_1, a_2,..., a_{n-1},x_n)$$ 
Once solved,
% all one-dimension subproblems, 
we back-propagate objective function values to the first-level $\phi_1(a_1)$ and continue searching from this level until the error bound is reached.

\subsubsection{Convergence Analysis}

We use mathematical induction to prove convergence for the multi-dimension case.
\begin{itemize}
	\item Base case: for all $ x\in \mathbb{R}$, $\lim_{i\to \infty}l_i = \inf_{x\in [a,b]} \of(x)$ and $\lim_{i\to \infty}(u_i - l_i) = 0$ hold.
	
	\item Inductive step: if, for all $ {x}\in \mathbb{R}^k$, $\lim_{i\to \infty}l_i = \inf_{{x}\in [a,b]^k} \of({x})$ and $\lim_{i\to \infty}(u_i - l_i) = 0$ are satisfied, then, for all $ {x}\in \mathbb{R}^{k+1}$, $\lim_{i\to \infty}l_i = \inf_{{x}\in [a,b]^{k+1}} \of({x})$ and $\lim_{i\to \infty}(u_i - l_i) = 0$ hold.
\end{itemize}
The base case (\ie~one-dimensional case) is already proved in Section~\ref{sec:onedimensional}. Now we prove the inductive step.

\begin{myproof}
	By the nested optimization scheme, we have
$$
	\min_{\mathbf{x} \in [a_i,b_i]^{k+1}}~~\of(\mathbf{x}) =\min_{x \in [a,b]}\Phi(x)
	$$
	$$
	\Phi(x) = \min_{\mathbf{y} \in [a_i,b_i]^k} \of(x,\mathbf{y})
$$
	Since $\min_{\mathbf{y} \in [a_i,b_i]^k} \of(x,\mathbf{y})$ is bounded by an interval error $\epsilon_{\mathbf{y}}$, assuming $\Phi^*(x)$ is the accurate global minimum, then we have
	$$\Phi^*(x)-\epsilon_{\mathbf{y}}\leq \Phi(x) \leq \Phi^*(x)+\epsilon_{\mathbf{y}}$$ 
	So the $k+1$-dimensional problem is reduced to the one-dimensional problem %decomposed as problem 
	$\min_{x \in [a,b]}\Phi(x)$. The difference from the real one-dimensional case is that evaluation of $\Phi(x)$ is not accurate but bounded by $|\Phi(x) - \Phi^*(x)|\leq \epsilon_{\mathbf{y}}, \forall x \in [a,b]$, where $\Phi^*(x)$ is the accurate function evaluation.
	
	Assuming that the minimal value obtained from our method is $\Phi^*_{min} = \min_{x\in [a,b]} \Phi^*(x)$ under accurate function evaluation, then the corresponding lower and upper bound sequences are $\{l^*_0, ..., l^*_i\}$ and $\{u^*_0,..., u^*_i\}$, respectively.
	
	For the inaccurate evaluation case, we assume $\Phi_{min} = \min_{x\in [a,b]} \Phi(x)$, and its lower and bound sequences are, respectively, $\{l_0, ..., l_i\}$ and $\{u_0, ..., u_i\}$. The termination criteria for both cases are $|u^*_i-l^*_i|\leq \epsilon_x$ and $|u_i-l_i|\leq \epsilon_x$, and $\phi^*$ represents the ideal global minimum. Then we have $\phi^* - \epsilon_x \leq l_i$. Assuming that $l^*_i \in [x_k,x_{k+1}]$ and $x_k, x_{k+1}$ are adjacent evaluation points, then due to the fact that $ l^*_i = \inf_{x\in [a,b]} H(x;\mathcal{Y}_{i}) $ we have
$$\phi^* - \epsilon_x \leq l^*_i = \dfrac{\Phi^*(x_k) + \Phi^*(x_{k+1})}{2} - \dfrac{L(x_{k+1} - x_k)}{2}$$
	Since $|\Phi(x_{i}) - \Phi^*(x_{i})|\leq \epsilon_{\mathbf{y}}, \forall i = k, k+1$,
% $$
% 	[\Phi^*(x_k) + \Phi^*(x_{k+1}) ]/2 \leq [\Phi (x_k) + \Phi (x_{k+1}) ]/2 + \epsilon_{\mathbf{y}}
% $$
we thus have
$$
	\phi^* - \epsilon_x \leq \dfrac{\Phi (x_k) + \Phi (x_{k+1}) }{2} + \epsilon_{\mathbf{y}} - \dfrac{ L(x_{k+1} - x_k)}{2}
$$
	Based on the search scheme, we know that 
	\begin{equation}l_i = \dfrac{\Phi(x_k) + \Phi(x_{k+1})}{2} - \dfrac{L(x_{k+1} - x_k)}{2}\end{equation}
	and thus we have $\phi^* - l_i \leq \epsilon_{\mathbf{y}} +\epsilon_x $. 
	
	Similarly, we can get
	\begin{equation}
	\phi^* + \epsilon_x \geq u^*_i = \inf_{y\in \mathcal{Y}_i } \Phi^*(y) \geq u_i -\epsilon_{\mathbf{y}}
\end{equation}
	so $u_i - \phi^* \leq \epsilon_x + \epsilon_{\mathbf{y}}$. By $\phi^* - l_i \leq \epsilon_{\mathbf{y}} +\epsilon_x$ and the termination criteria $u_i - l_i \leq \epsilon_x$, we have $l_i - \epsilon_{\mathbf{y}} \leq \phi^* \leq u_i + \epsilon_{\mathbf{y}}$, \ie~the accurate global minimum is also bounded.
\end{myproof}

The proof indicates that the overall error bound of the nested scheme only increases linearly w.r.t. the bounds in the one-dimensional case. 
Moreover, an adaptive approach can be applied to optimise its performance without compromising convergence. The
key observation is to relax the strict subordination inherent in the nested scheme and 
simultaneously consider all the univariate subproblems arising in the course of
multidimensional
optimization.
For all the generated subproblems that are active, a numerical measure is applied.
Then an iteration of the multidimensional optimization consists in choosing the subproblem
with maximal measurement and carrying out a new trial within this subproblem.
The measure is defined to be the maximal interval characteristics generated by the one-dimensional optimisation algorithm. 

%\wenjie{
\section{Proof of NP-completeness}

In this section, we prove the NP-completeness of our generic reachability problem. 

\subsection{Upper Bound}

First of all, we show that the one-dimension optimisation case is linear with respect to the error bound $\epsilon$. 
    As shown in Figure~\ref{fig-1}, we have that
    
    $$
    \of(y_{k-1}) - z^* = K(y_i - y_{k-1})
    $$
    $$
    \of(y_{k}) - z^* = K(y_k - y_{i})
    $$
    
    So we have
    $$
    \of(y_k) - \of(y_{k-1}) = K(y_k - y_{i}) - K(y_i - y_{k-1})
    $$
    Moreover, we have
    $$
    \begin{array}{l}
    \of(y_k) + \of(y_i) - K(y_k - y_i) 
    - \dfrac{\of(y_k) + \of(y_{k-1})}{2} + \\ 
    \hfill\dfrac{K(y_k - y_{k-1})}{2} = \of(y_i)
    \end{array}
    $$
    Based on Equation~(\ref{eqn-17}) and (\ref{eqn-19}), we have $$2z_i - z^* = \of(y_i)$$
    Therefore, we have
    $z_{i} - z^* = \dfrac{w(y_{i}) - z^*}{2}$.
Now, since we have that $w(y_{i})\geq u^*$ and $z^*\leq l^*$, and $u^*-l^*>\epsilon$ before convergence, we have
$$
z_{i} - z^* > \dfrac{1}{2}\epsilon
$$
Therefore, the improvement for each iteration is of linear with respect to the error bound $\epsilon$, which means that the optimisation procedure will converge in linear time with respect to the size of region $[a,b]$.

For the multiple dimensional case, we notice that, in Equation (\ref{equ-16}), to reach the global optimum, not all the dimensions $x_i$ for $i\in [1..n]$ need to be changed and the ordering between dimensions matter. 
Therefore, we can have a non-deterministic algorithm which guesses a subset of dimensions together with their ordering. These are dimensions that need to be changed to lead from the original input to the global optimum. This guess can be done in polynomial time. 

Then, we can apply the one-dimensional optimisation algorithm backward from the last dimension to the first dimension. Because of the polynomial time convergence of the one-dimensional case, this procedure can be completed in polynomial time. 

Therefore, the entire procedure can be completed in polynomial time with a nondeterministic algorithm, i.e., in NP. 

\subsection{Lower Bound}

We have a reduction from the 3-SAT problem, which is known to be NP-complete. A 3-SAT Boolean formula $\varphi$ is of the form $c_1 \land ... \land c_m$, where each clause $c_i$ is of the form $l_{i1}\lor l_{i2}\lor l_{i3}$. Each literal $l_{ij}$, for $1\leq i\leq m$ and $1\leq j\leq 3$, is a variable $v$ or its negation $\neg v$,  such that $v\in \{v_1,...,v_n\}$. 
The 3-SAT problem is to decide the existence of a truth-value assignment to the boolean variables $V=\{v_1,...,v_n\}$ such that the formula $\varphi$ is True. 

\subsubsection{Construction of DNN}

We let $var_{ij}\in V$ and $sgn_{ij}\in\{p,n\}$ be the variable and the sign of the literal $l_{ij}$, respectively.
Given a formula $\varphi$, we construct a DNN $f$ which implements a classification problem. The DNN has four layers $\{L_i\}_{i\in [1..4]}$, within which $L_2$, $L_3$ are hidden layers, $L_1$ is the input layer, and $L_4$ is the output layer. It has $n$ input neurons and $m$ output neurons. 

\paragraph{Input Layer} 

The input layer $L_1$ has $n$ neurons, each of which represents a variable in $V$. 

\paragraph{First Hidden Layer -- Fully Connected with ReLU}

The hidden layer $L_2$ has $2n$ neurons, such that every two neurons correspond to a variable in $V$. Given a variable $v_i$, we write $pv_i$ and $nv_i$ to denote the two neurons for $v_i$ such that 
\begin{equation}
\begin{array}{l}
pv_i = ReLU(v_i) \\
nv_i = ReLU(-1 * v_i)
\end{array}
\end{equation}
It is noted that, the above functions can be implemented as a fully connected function, by letting the coefficients from  variables other than $v_i$ be 0. 

\paragraph{Second Hidden Layer -- Fully Connected with ReLU} 

The hidden layer $L_3$ has $m$ neurons, each of which represents a clause. Let $cv_i$ be the neuron representing the clause $c_i$. Then for a clause $c_i= l_{i1} \lor l_{i2} \lor l_{i3}$, we have  
\begin{equation}
cv_i = ReLU(xv_{i1}+xv_{i2}+xv_{i3}) 
\end{equation}
where for $k\in\{1,2,3\}$, we have 
$$
xv_{ik} =  \left\{
\begin{array}{cc}
nv_j     & \text{ if } sgn_{ik} = n \text{ and } var_{ik} = v_j \text{ for some }j  \\
pv_j     & \text{ if } sgn_{ik} = p \text{ and } var_{ik} = v_j \text{ for some }j 
\end{array}
\right.
$$
Intuitively, $cv_i$ takes either a positive value or zero. For the latter case, none of the three literals are satisfied. 

\paragraph{Output Layer  -- Fully Connected Without ReLU}

The output layer $L_4$ has $m$ neurons, each of which represents a clause. Let $ncv_i$ be the neuron representing the clause $c_i$. Then we have 
\begin{equation}
ncv_i = -1 * cv_i
\end{equation}
Intuitively, this layer simply negates all the values from the previous layer. 

\subsubsection{Statistical Evaluation Function $o$}
 
After the output, we let $o$ be the following function
\begin{equation}
ov = \max \{ ncv_i ~|~ i \in [1..m]\}
\end{equation}
That is, $o$ gets the maximal value of all the outputs. Because $cv_i \geq 0$ and $ncv_i \leq 0$, we have that $ov\leq 0$.

\subsubsection{Reduction}

First, we show that for any point $x \in \mathbb{R}^n$, there exists a point $x^0 \in \{-1,1\}^n$ such that, $\of(x) \neq 0 $ if and only if $\of(x^0)\neq 0$. Recall that $\of$ is a concatenation of the network $f$ with $o$, i.e., $\of = o \cdot f$. For any input dimension $i \in [1..n]$, we let 
$$
x^0(i)= \left\{
\begin{array}{cc}
    1  & \text{ if } x(i) \geq 0 \\
    -1 & \text{ if } x(i) < 0
\end{array}
\right.
$$
Therefore, by construction, we have that $cv_i = 0$ if and only if $cv_i^0 = 0$. After passing through $L_4$ and the function $o$, we have that  $ov = 0$ if and only $ov^0 = 0$. This is equivalent to $\of(x) \neq 0 $ if and only if $\of(x^0)\neq 0$. 

Then, for every truth-assignment $a:V \rightarrow \{True,False\}$, we  associate with it a point $x \in \{-1,1\}^n$ by letting $x(i) = 1$ if $a(i)= True$ and $x(i) = -1$ if $a(i)=False$.

Finally, we show that the formula $\varphi$ is satisfiable if and only if the function $\of=o\cdot f$ cannot reach value 0. This is done by only considering those points in $\{-1,1\}^n$. 

$(\Rightarrow)$ If $\varphi$ is satisfiable then by construction, in the second hidden layer, $cv_i > 0$ for all $i\in [1..m]$. Therefore, in the output layer, we have $ncv_i < 0$ for all $i\in [1..m]$. Finally, with the function $o$, we have $ov < 0$, i.e., the function $w$ cannot reach value 0. 

$(\Leftarrow)$ We prove by contradiction. If $\varphi$ is unsatisfiable, then there must exist a clause which is unsatisfiable. Then by construction, we have $cv_i=0$ for some $i\in [1..m]$. This results in $ncv_i=0$ and $ov=0$, which contradicts with the hypothesis that $w$ cannot reach 0.

\begin{figure}[t]
	\centering
	\includegraphics[width=0.8\linewidth]{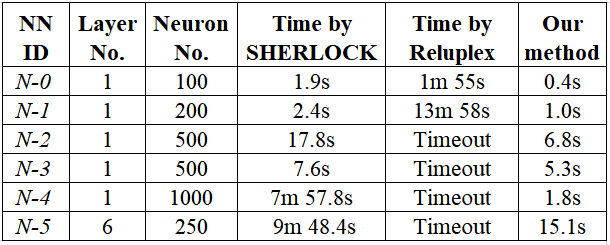}
	\caption{Comparison with SHERLOCK and Reluplex}
	\label{fig-2}
\end{figure}

\begin{figure}[ht]
	\centering
	\includegraphics[width=1\linewidth]{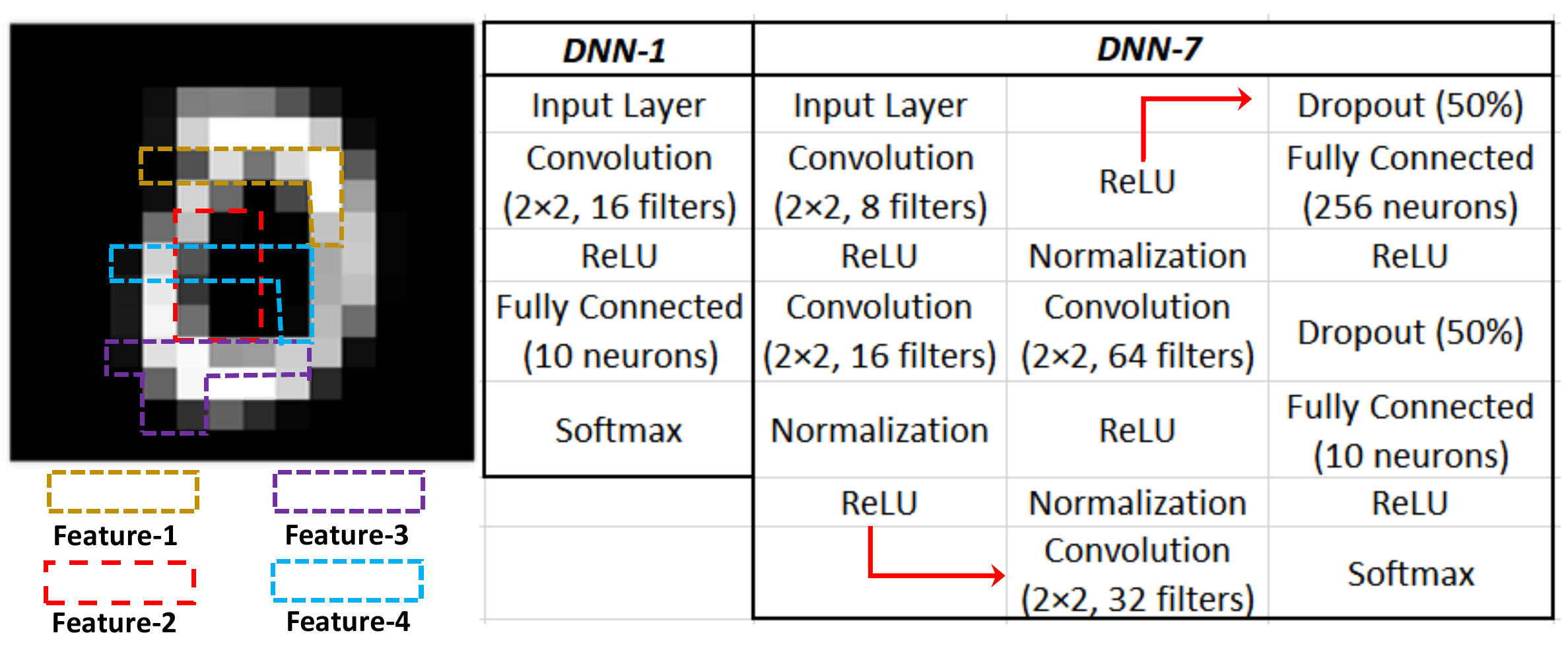}
	\caption{The four features and the architecture of DNN-1 and DNN-7}
	\label{fig-3(a)}
\end{figure}
\begin{figure}[ht]
	\centering
	\includegraphics[width=1.01\linewidth]{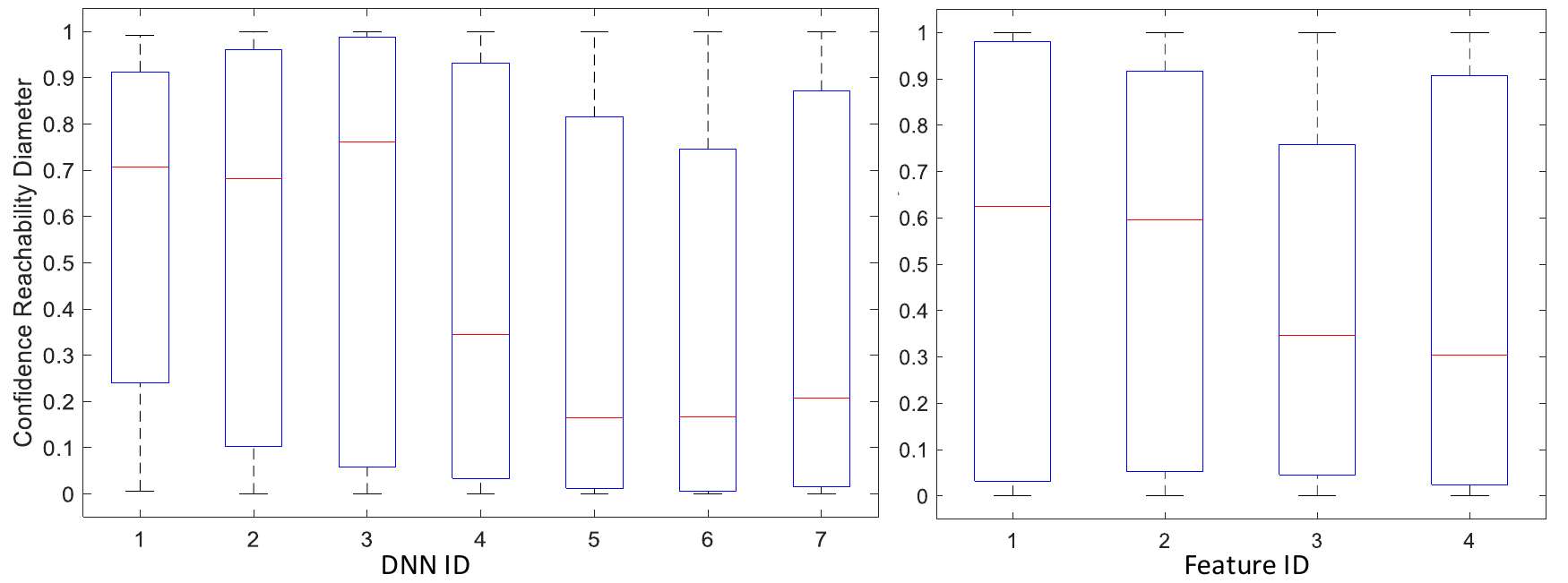}
	\caption{Left: boxplots of confidence reachability diameters for 7 DNNs, based on $4\times20$ analyses of each DNN. Right: boxplot of confidence reachability diameters for 4 features, based on $7\times20$ analyses of each feature. The red line represents the median value: a lower value indicates a more robust model or feature.}
	\label{fig-3(b)}
\end{figure}
\section{Experiments}
\subsection{Comparison with State-of-the-art Methods}

Two methods are chosen as baseline methods in this paper:
\begin{itemize}
    \item Reluplex~\cite{katz2017reluplex}: an SMT-based method for solving queries on DNNs with ReLU activations; we apply a bisection scheme to compute an interval until an error is reached
    \item SHERLOCK~\cite{dutta2017output}: a MILP-based method dedicated to output range analysis on DNNs with ReLU activations.
\end{itemize}

Our software is implemented in Matlab 2018a, running on a notebook computer with i7-7700HQ CPU and 16GB RAM. Since Reluplex and SHERLOCK (not open-sourced) are designed on different software platforms, we take their experimental results from~\cite{dutta2017output}, whose experimental environment is a Linux workstation with 63GB RAM and 23-Cores CPU (more powerful than ours) and $\epsilon=0.01$.
Following the experimental setup in~\cite{dutta2017output}, we use their data (2-input and 1-output functions) to train six neural networks with various numbers and types of layers and neurons. The input subspace is $X' = [0,10]^2$. 

The comparison results are given in Fig.~\ref{fig-2}. 
They show that, while the performance of both Reluplex and SHERLOCK is considerably affected by the increase in the number of neurons and layers, our method is not. For the six benchmark neural networks, our average computation time is around $5s$, 36 fold improvement over SHERLOCK and nearly 100 fold improvement over Reluplex (excluding timeouts). We note that our method is running on a notebook PC, which is significantly less powerful than the 23-core CPU stations used for SHERLOCK and Reluplex.

\begin{figure}[t]
	\centering
	\includegraphics[width=0.7\linewidth]{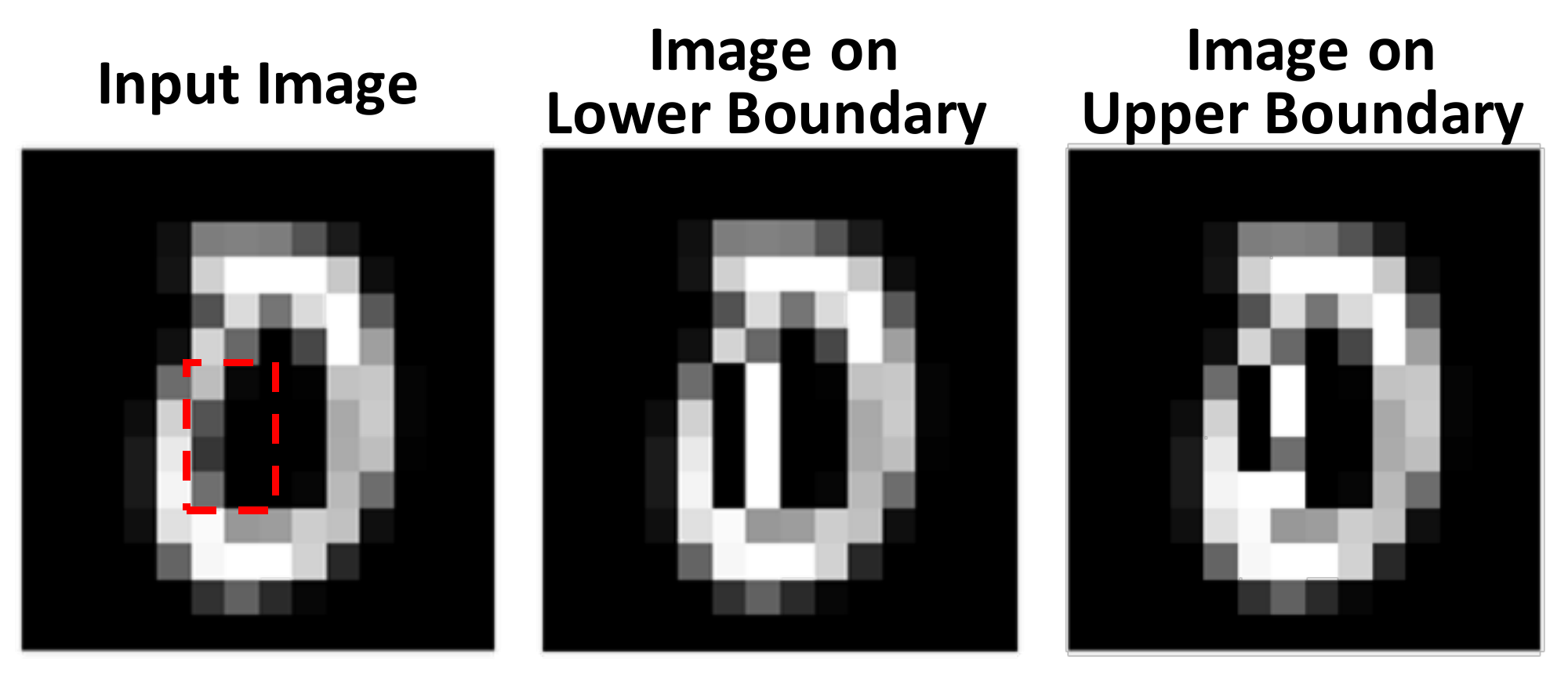}
	\caption{Left: an original image (logit is 11.806, confidence of output being `0' is 99.95\%), where area marked by dashed line is the feature. Middle: an image on the confidence lower bound. Right: an image on the confidence upper bound; for the output label `0', the feature's output range is $[74.36\%,99.98\%]$, and logit reachability is $[7.007, 13.403]$}
	\label{fig-4(a)}
\end{figure}
\begin{figure}
	\centering
	\includegraphics[width=0.8\linewidth]{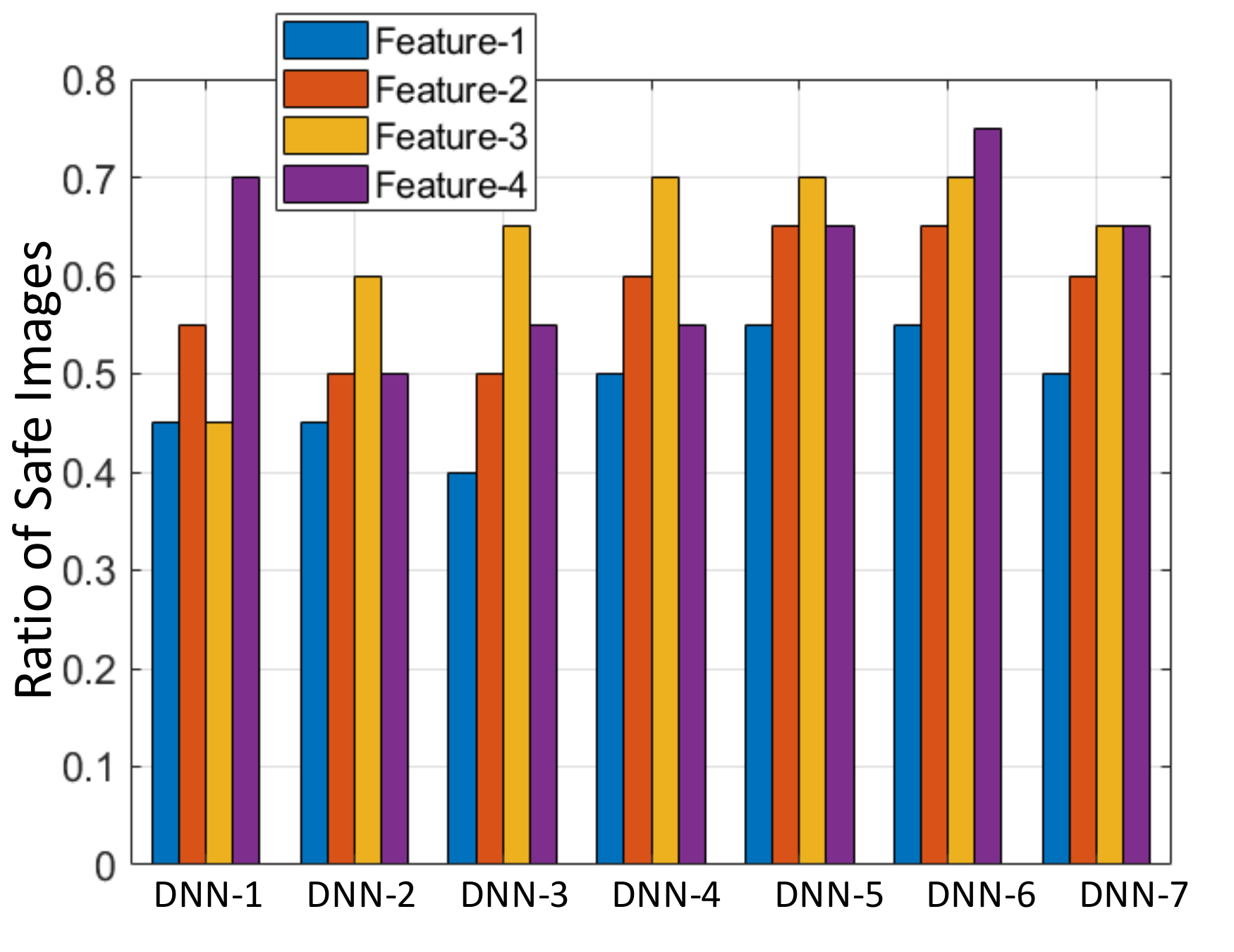}
	\caption{Ratios of safe images for 7 DNNs and 4 features}
	\label{fig-4(b)}
\end{figure}

\begin{figure}[ht]
	\centering
	\includegraphics[width=1\linewidth]{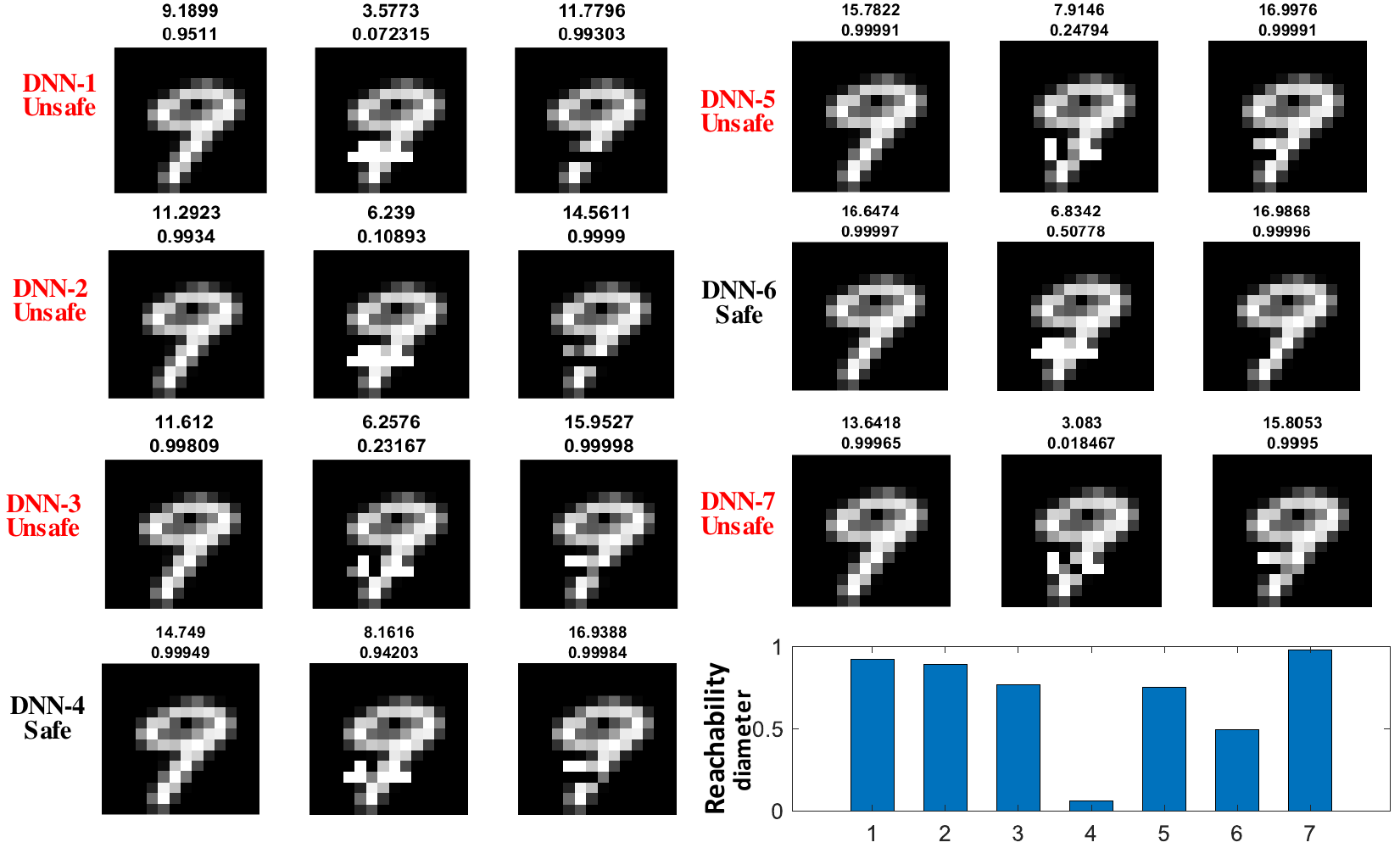}
	\caption{A detailed example comparing the safety and robustness of DNNs for image '9' and Feature-3: the top number in the caption of each figure is logit and the bottom one is confidence; the unsafe cases are all misclassified as `8'; the last bar chart shows their confidence reachability diameters.}
	\label{fig-4(c)}
\end{figure}

\begin{sidewaysfigure}[h]
		\begin{center}
			\includegraphics[width=1\linewidth]{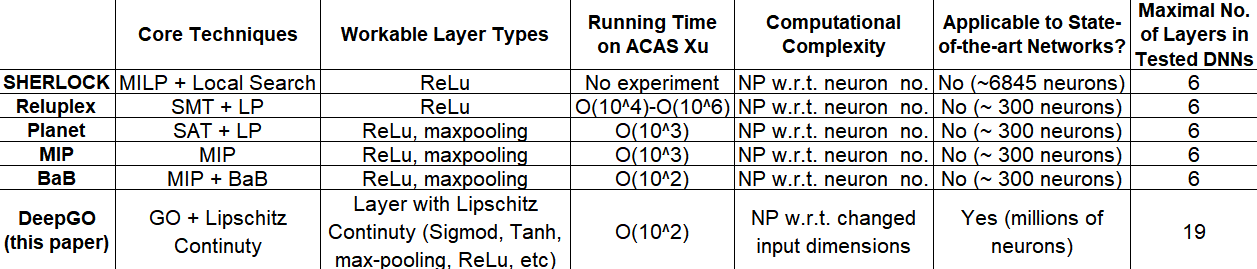}
	\caption{A high-level comparison with state-of-the-art methods: SHERLOCK~\cite{dutta2017output}, Reluplex~\cite{katz2017reluplex}, Planet~\cite{ehlers2017formal}, MIP~\cite{CNR2017,LM2017} and BaB~\cite{bunel2017piecewise}.}
	\label{figr-1}
		\end{center}
\end{sidewaysfigure}

\subsection{Safety and Robustness Verification by Reachability Analysis}

We use our tool to conduct logit and output range analysis.
%for verifying the safety of deep neural networks, and compare robustness of different networks and input subspaces. 
Seven convolutional neural networks, represented as DNN-1,...,DNN-7, were trained on the MNIST dataset. Images are resized into $14\times 14$ to enforce that a DNN with deeper layers tends to over-fit. The networks have different layer types, including ReLu, dropout and normalization, and the number of layers ranges from $5$ to $19$. Testing accuracies range from $95\%$ to $99\%$, and $\epsilon=0.05$ is used in our experiments.

We randomly choose 20 images (2 images per label) and manually choose 4 features
%\footnote{We also conduct experiments in which features are selected by object detection techniques such as SIFT \cite{Lowe1999}. We cannot include the results for space limit, but the conclusions are similar. } 
such that each feature contains 8 pixels, i.e., $X'=[0,1]^8$. Fig.~\ref{fig-3(a)} illustrates the four features and the architecture of two DNNs with the shallowest and deepest layers, \ie~DNN-1 and DNN-7. 

\vspace{2mm}

\noindent{\bf Safety Verification}
%
%Then we conduct the reachability analysis for those features. 
Fig.~\ref{fig-4(a)} shows an example: for DNN-1, Feature-4 is \emph{guaranteed to be safe} with respect to the image $x$ and the input subspace $X'$.
%to any adversarial perturbation. 
Specifically, the reachability interval is $R(\Pi_0,X',\epsilon) = [74.36\%,99.98\%]$, which means that $l(\Pi_0,X',\epsilon)=74.36\%$. By this, we have $u(\oplus_{-0},X',\epsilon) \leq (1-0.7436) < 0.7436 = l(\Pi_0,X',\epsilon)$. Then, by Theorem \ref{thm:safety}, we have 
$\Safe(\text{DNN-1},x,X')$. Intuitively, no matter how we manipulate this feature, the worst case is to reduce the confidence of output being `0' from 99.95\% (its original confidence probability) to 74.36\%. 

\vspace{2mm}

\noindent{\bf Statistical Comparison of Safety}
Fig.~\ref{fig-4(b)} compares the ratios of safe images for different DNNs and features. It shows that: \one~no DNN is 100\% safe on those features: DNN-6 is the safest one and DNN-1, DNN-2 and DNN-3 are less safe, which means a DNN with well chosen layers are safer than those DNNs with very shallow or deeper layers; and \two~the safety performance of different DNNs is consistent for the same feature, which suggests that the feature matters -- some features are easily perturbed to yield adversarial examples, e.g., Feature-1 and Feature-2.

\vspace{1mm}

\noindent{\bf Statistical Comparison of Robustness}
Fig.~\ref{fig-3(b)} compares the robustness of networks and features with two boxplots over the reachability diameters, where the function $o$ is $\Pi_j$ for a suitable $j$. We can see that DNN-6 and DNN-5 are the two most robust, while DNN-1, DNN-2 and DNN-3 are less robust. Moreover, Feature-1 and Feature-2 are less robust than Feature-3 and Feature-4. 

%Together with the above, we can see that 
We have thus demonstrated that
reachability analysis with our tool can be used to quantify the safety and robustness of deep learning models. In the following, we perform a comparison of networks over a fixed feature. 

\vspace{2mm}

\noindent{\bf Safety Comparison of Networks}
By Fig.~\ref{fig-4(c)}, DNN-4 and DNN-6 are guaranteed to be safe w.r.t.\ the subspace defined by Feature-3.
%, particularly DNN-4 whose reachability range is $[94.2\%, 100\%]$, which is very robust. 
Moreover, the output range of DNN-7 is $[1.8\%, 100.0\%]$, %It essentially indicates that, given confidence (of being classified as '9') in $[1.8\%, 100.0\%]$, 
which means that we can generate adversarial images by only perturbing this feature, among which the worst one is as shown in the figure with a confidence 1.8\%. 
Thus, reachability analysis not only enables qualitative safety verification (\ie~safe or not safe), but also allows benchmarking of safety of different deep learning models in a principled, quantitive manner (\ie~how safe) by quantifying the `worst' adversarial example. Moreover, compared to retraining the model with `regular' adversarial images, these `worst' adversarial images are more effective in improving the robustness of DNNs~\cite{kolter2017provable}.

\vspace{2mm}

\noindent{\bf Robustness Comparison of Networks}
The bar chart in Fig.~\ref{fig-4(c)} shows 
%those lower boundary and upper boundary images for image '9' and feature-3 of all DNNs. We can see 
 the reachability diameters of the networks over Feature-3, where the function $o$ is $\Pi_j$. DNN-4 is the most robust one, and its output range is $[94.2\%, 100\%]$. 
 %This experiments which can potentially be adopted to quantitatively compare the robustness of different DNNs or different input subspace/features for the same DNNs (a smaller reachability diameter means better robustness). 

\subsection{A Comprehensive Comparison with the State-of-the-arts}\label{sec:technicalComparison}

This section presents a comprehensive, high-level comparison of our method with several existing approaches that have been used for either range analysis or verification of DNNs, including SHERLOCK~\cite{dutta2017output}, Reluplex~\cite{katz2017reluplex}, Planet~\cite{ehlers2017formal}, MIP~\cite{CNR2017,LM2017} and BaB~\cite{bunel2017piecewise}, as shown in Fig.~\ref{figr-1}. We investigate these approaches from the following seven aspects:  
\begin{enumerate}
    \item core techniques,
    \item workable layer types, 
    \item running time on ACAS Xu, 
    \item computational complexity,
    \item applicable to state-of-the-art networks,
    \item input constraints, and 
    \item maximum number of layers in tested DNNs. 
\end{enumerate}
We are incomparable to approaches based on exhaustive search (such as DLV~\cite{HKWW2017} and SafeCV~\cite{WHK2017}) because we have a different way of expressing guarantees. 

\vspace{2mm}

\noindent{\bf Core Techniques } Most existing approaches (SHERLOCK, Reluplex, Planet, MIP) are based on reduction to constraint solving, except for BaB which mixes constraint solving with local search. On the other hand, our method is based on global optimization and assumes Lipschitz continuity of the networks. As indicated in Section 3, %of the paper, 
all known layers used in classification tasks are Lipschitz continuous. 

\vspace{2mm}

\noindent{\bf Workable Layer Types }
While we are able to work with all known layers used in classification tasks because they are Lipschitz continuous (proved in Section 3 of the paper), Planet, MIP and BaB can only work with Relu and Maxpooling, and  SHERLOCK and Reluplex can only work with Relu. 

\vspace{2mm}

\noindent{\bf Running Time on ACAS-Xu Network } 
We collect running time data from~\cite{bunel2017piecewise} on the ACAS-Xu network, and find that our approach has similar performance to BaB, and better than the others. No experiments for SHERLOCK are available. We reiterate that, compared to their experimental platform (Desktop PC with i7-5930K CPU, 32GB RAM), ours is less powerful (Laptop PC with i7-7700HQ CPU, 16GB RAM). 
We emphasise that, although our approach performs well on this network, the actual strength of our approach is not the running time on small networks such as ACAS-Xu, but  the ability to work with large-scale networks (such as those shown in Section 6.2).

\vspace{2mm}

\noindent{\bf Computational Complexity }
While all the mentioned approaches are in the same complexity class, NP, the complexity of our method is with respect to the number of input dimensions to be changed, as opposed to %. This is compared with the other approaches which are with respect to 
the number of hidden neurons. It is known that the number of hidden neurons is much larger than the number of input dimensions, \eg~there are nearly $6.5\times10^6$ neurons in AlexNet.

\vspace{2mm}

\noindent{\bf Applicable to State-of-the-art Networks }
We are able to work with state-of-the-art networks with millions of neurons. However, the other tools (Reluplex, Planet, MIP, BaB) can only work with hundreds of neurons. SHERLOCK can work with thousands of neurons thanks to its interleaving of MILP with local search.  

\vspace{2mm}

\noindent{\bf Maximum Number of Layers in Tested DNNs }
We have validated our method on networks with 19 layers, whereas the other approaches are validated on up to 6 layers.

In summary, the key advantages of our approach are as follows: \one~the ability to work with large-scale state-of-the-art networks; \two~lower computational complexity, i.e., NP-completeness with respect to the input dimensions to be changed, instead of the number of hidden neurons; and \three~the wide range of types of layers that can be handled.

%}

\section{Conclusion}

We propose, design and implement a reachability analysis tool for deep neural networks, which has provable guarantees and can be applied to neural networks with deep layers and nonlinear activation functions. The experiments demonstrate that our tool can be utilized to verify the safety of deep neural networks and quantitatively compare their robustness. We envision that this work marks an important step towards a practical, guaranteed safety verification for DNNs. Future work includes parallelizing this method in GPUs to improve its scalability on large-scale models trained on ImageNet, and a generalisation %of this method to work with 
to other deep learning models such as RNNs and deep reinforcement learning.

\vspace{1mm}

\section{Acknowledgements}
WR and MK are supported by the EPSRC Programme Grant
on Mobile Autonomy (EP/M019918/1). XH   acknowledges NVIDIA Corporation for its support with the donation of the Titan Xp GPU, and is partially supported by NSFC (no. 61772232).

\bibliographystyle{IEEEtran}
\bibliography{ijcai18_2}

\end{document}